\documentclass[pdflatex,sn-mathphys-num]{sn-jnl}
\usepackage{graphicx}%
\usepackage{multirow}%
\usepackage{amsmath,amssymb,amsfonts}%
\usepackage{amsthm}%
\usepackage{mathrsfs}%
\usepackage[title]{appendix}%
\usepackage{xcolor}%
\usepackage{textcomp}%
\usepackage{manyfoot}%
\usepackage{booktabs}%
\usepackage{algorithm}%
\usepackage{algorithmicx}%
\usepackage{algpseudocode}%
\usepackage{listings}%
\usepackage{bibunits}
\theoremstyle{thmstyleone}%
\theoremstyle{thmstyletwo}%
\usepackage{amsmath}
\usepackage{amsmath}
\DeclareMathOperator*{\argmax}{arg\,max}

\newcommand{\argminD}{\arg\!\min} 

\theoremstyle{thmstylethree}%

\begin{document}

\addtocontents{toc}{\protect\setcounter{tocdepth}{0}}

\title[Article Title]{Generative assimilation and prediction for weather and climate}


\author[1,2,3,4]{\fnm{Shangshang} \sur{Yang}}
\equalcont{These authors contributed equally to this work.}

\author[1]{\fnm{Congyi} \sur{Nai}}
\equalcont{These authors contributed equally to this work.}

\author[1,5]{\fnm{Xinyan} \sur{Liu}}

\author[1]{\fnm{Weidong} \sur{Li}}

\author[1]{\fnm{Jie} \sur{Chao}}

\author[1,6]{\fnm{Jingnan} \sur{Wang}}

\author[7,8]{\fnm{Leyi} \sur{Wang}}

\author[1]{\fnm{Xichen} \sur{Li}}

\author[1]{\fnm{Xi} \sur{Chen}}

\author[9,10]{\fnm{Bo} \sur{Lu}}


\author[1]{\fnm{Ziniu} \sur{Xiao}}

\author[3,4]{\fnm{Niklas} \sur{Boers}}

\author*[2]{\fnm{Huiling} \sur{Yuan}}

\author*[1]{\fnm{Baoxiang} \sur{Pan}}\email{panbaoxiang@lasg.iap.ac.cn}

\affil*[1]{\orgdiv{Institute of Atmospheric Physics}, \orgname{Chinese Academy of Sciences}, \city{Beijing}, \country{China}}

\affil*[2]{\orgdiv{School of Atmospheric Sciences and Key Laboratory of Mesoscale Severe Weather/Ministry of Education}, \orgname{Nanjing University}, \city{Nanjing}, \country{China}}

\affil[3]{\orgdiv{Earth System Modelling, School of Engineering and Design}, \orgname{Technical University of Munich}, \city{Munich}, \country{Germany}}

\affil[4]{\orgdiv{Complexity Science}, \orgname{Potsdam Institute for Climate Impact Research}, \city{Potsdam}, \country{Germany}}

\affil[5]{\orgname{Chinese Academy of Meteorological Sciences}, \city{Beijing}, \country{China}}

\affil[6]{\orgdiv{College of Meteorology and Oceanography}, \orgname{National University of Defense Technology}, \city{Changsha}, \country{China}}

\affil[7]{\orgdiv{School of Mathematical Sciences}, \orgname{Peking University}, \city{Beijing}, \country{China}}

\affil[8]{\orgdiv{Chongqing Research Institute of Big Data}, \orgname{Peking University}, \city{Chongqing}, \country{China}}

\affil[9]{\orgdiv{China Meteorological Administration Key Laboratory for Climate Prediction Studies}, \orgname{National Climate Center}, \city{Beijing}, \country{China}}

\affil[10]{\orgname{Xiong’an Institute of Meteorological Artificial Intelligence}, \city{Xiong’an}, \country{China}}






\abstract{Machine learning models have shown great success in predicting weather up to two weeks ahead, outperforming process-based benchmarks \cite{bi2023accurate, lam2023learning, kochkov2024neural, price2024probabilistic}. However, existing approaches mostly focus on the prediction task, and do not incorporate the necessary data assimilation. Moreover, these models suffer from error accumulation in long roll-outs, limiting their applicability to seasonal predictions \cite{chen2024machine} or climate projections \cite{watson2022climatebench}. 
Here, we introduce Generative Assimilation and Prediction (GAP), a unified deep generative framework for assimilation and prediction of both weather and climate. 
By learning to quantify the probabilistic distribution of atmospheric states under observational, predictive, and external forcing constraints, 
GAP excels in a broad range of weather-climate related tasks, including data assimilation, seamless prediction, and climate simulation. In particular, GAP is competitive with state-of-the-art ensemble assimilation, probabilistic weather forecast and seasonal prediction, yields stable millennial simulations, and reproduces climate variability from daily to decadal time scales.
}

\maketitle

\subsection*{Introduction}\label{Main}

Reliable modeling of Earth's chaotic atmosphere is fundamental to predicting both day-to-day weather and long-term climate change. 
Recent decades have seen remarkable advances in our understanding and simulation of atmospheric dynamics \cite{bauer2015quiet}. These improvements have enhanced forecasting through two key processes: 
data assimilation (Fig.~\ref{fig:schematic}A left), 
which combines observations with models to infer current atmospheric states \cite{evensen2022data}, and prediction (Fig.~\ref{fig:schematic}A right), 
which projects the initial state estimates forward in time, 
until chaos inevitably blurs our view of the future \cite{palmer2005probabilistic}.
At the core of both processes lies a fundamental challenge — how to accurately represent and evolve the probabilities of possible atmospheric states, given sparse observations, imperfect models, and complex natural and anthropogenic forcings \cite{eyring2016overview,palmer2019scientific}. 
Successfully addressing this challenge would bridge a critical gap between weather forecasting and climate projection \cite{white2017potential}, while  connecting our understanding of individual atmospheric processes \cite{manabe1967thermal} to the climate system's overall behavior \cite{von1988principal}. 
Despite significant advances in weather and climate modeling, accurately capturing these probability distributions remains difficult, due to model biases, high dimensionality, and computational constraints.

Machine learning (ML) has emerged as a transformative avenue for weather forecast \cite{bi2023accurate, lam2023learning, kochkov2024neural, price2024probabilistic}.  By identifying and leveraging analogous patterns in historical weather trajectories, ML models can make direct temporal predictions, bypassing step-by-step resolving  of geophysical fluid dynamics.
These approaches have achieved competitive forecasts up to two weeks ahead with unprecedented computational efficiency.
However, current ML methods often focus narrowly on minimizing forecast errors without addressing other essential aspects of the forecasting process \cite{hakim2024dynamical, rackow2024robustness, bracco2024machine, rasp2024weatherbench}. 
These models usually 
rely on initial conditions from separate data assimilation systems, and struggle to incorporate new observations.
Also, they tend to overlook forecast uncertainties, and underestimate extreme events. Furthermore, while progress has been made in extending ML approaches to longer time ranges \cite{kochkov2024neural, watt2024ace2, cresswell2024deep, wang2024coupled, chen2024machine}, ML models have not achieved the stability needed for seasonal prediction, nor demonstrated the ability to reproduce key patterns of climate variability --- from regional weather patterns to (multi-)decadal variability and global climate phenomena. These limitations
underscore the necessity to rethink how ML can address the core challenges in weather and climate modeling.

The fundamental pursuit of climate science extends beyond efficient prediction -- we 
seek a comprehensive understanding of how the climate system 
evolves as a chaotic, high-dimensional, multi-scale system under various forcings \cite{palmer2016personal,eyring2016overview,palmer2019scientific,ghil2020hilbert}. 
Despite its complexity, atmospheric dynamics follow structured patterns across multiple scales. 
These patterns create a characteristic phase space -- the set of all possible atmospheric states and their probabilities of occurrence. This phase space distribution, also known as the climatological distribution \cite{palmer2016personal,ghil2020hilbert, ghil2020physics,pan2021learning}, represents the system's long-term behavior and encodes both its natural variability and its response to forcing.

Modern deep generative models 
\cite{sohl2015deep,ho2020denoising,song2020score} 
offer a powerful approach to approximate this phase space structure.
While previous applications of generative models to weather forecast have shown great promises \cite{ravuri2021skilful, hess2022physically, price2024probabilistic, cachay2024probabilistic, nai2024reliable},
they have largely followed a supervised learning paradigm that makes predictions conditional on complete input states. This creates two fundamental limitations:
First, these models can only function when their required initial conditions are fully available --- a major constraint given the sparse and irregular nature of atmospheric observations. Second, by focusing on predicting specific weather phenomena rather than modeling the complete distribution of atmospheric states, they cannot capture the full range of weather-climate interactions and multi-scale variability. A unified framework is thus urgently needed, which can operate with varying levels of observational information while preserving the intrinsic atmospheric dynamics that connect weather patterns to climate statistics.

We demonstrate  that a probabilistic diffusion model \cite{sohl2015deep,ho2020denoising,song2020score} can accurately capture the phase space distribution of Earth's atmospheric states, providing foundation for both inferring current state from limited observations, and predicting future state evolution across time scales. We introduce the Generative Assimilation and Prediction (GAP) methodology, which learns this distribution from historical climate records,  and adapts it to different tasks by incorporating observational, predictive, and external forcing constraints in a ``plug-and-play'', approximate Bayesian computation manner (Fig.~\ref{fig:schematic}B).
GAP demonstrates remarkable capabilities across multiple challenges in weather and climate prediction. In data assimilation, it outperforms leading operational systems \cite{bonavita2016evolution}  using less than 1\% of typical observations to accurately infer atmospheric states. For weather forecast, it efficiently provides uncertainty-aware predictions that match or exceed those from state-of-the-art process-based \cite{buizza2018development} and data-driven ML models \cite{bi2023accurate,lam2023learning}, or hybrid combinations of both \cite{kochkov2024neural}. For prediction at seasonal timescales, GAP demonstrates considerable potential: even with simple constraints, it surpasses current operational systems \cite{johnson2019seas5} in three-month forecasts. Perhaps most remarkably, GAP can run stable climate simulations over centuries while adapting to external forcing, a capability that has eluded previous machine learning approaches.


\subsection*{Generative Assimilation and Prediction}

\begin{figure}[H]
    \centering
    \includegraphics[width=1\linewidth]{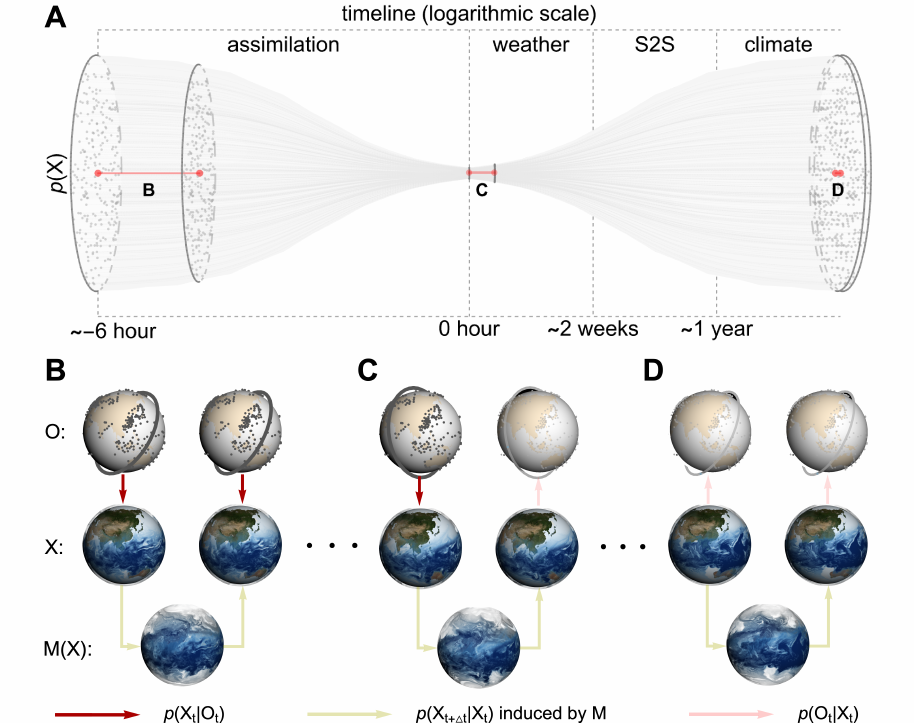}
    \caption{Schematic of the generative assimilation and prediction (GAP) framework. \textbf{A}, Evolution of atmospheric state probability distributions, illustrating the Probabilistic Assimilation and Prediction (PAP) ideology. A logarithmic timeline is adopted to 
    highlight the duality between assimilation and prediction that we aim to unify. During assimilation (left), observations progressively constrain the climatological distribution, narrowing possible states. During prediction (right), uncertainty evolves through distinct phases: weather forecasting (up to around two weeks) shows growing uncertainty until deterministic predictions lose utility; seasonal prediction (up to around 1-2 years) maintains predictability below climatological levels through low-frequency climate process constraints; and climate prediction (beyond 1-2 years) shows complete dissipation of initial condition information while external forcings modify the underlying climatological distribution. 
    \textbf{B-D}, Practical implementation of GAP.
    \textbf{B}, Assimilation phase implementation:
    given state estimates sampled from $p(\mathbf{X}_t|\mathbf{X}_{t-\Delta t},\mathbf{O}_t)$ -- the probability distribution of the current state $\mathbf{X}_t$ conditioned on previous state $\mathbf{X}_{t-\Delta t}$ (or climatological distribution if $\mathbf{X}_{t-\Delta t}$ were unavailable) and current observations $\mathbf{O}_t$ -- we use a forecasting model $\mathbf{M}$ to generate preliminary predictions, which together with observations $\mathbf{O}_{t+\Delta t}$ guide subsequent probabilistic state estimation for $p(\mathbf{X}_{t+\Delta t}|\mathbf{X}_{t},\mathbf{O}_{t+\Delta t})$. 
    \textbf{C}, Transition point marking the end of assimilation and beginning of prediction, where forecasts proceed without observational input while yielding predicted observations via $p(\mathbf{O}|\mathbf{X})$. \textbf{D}, Prediction phase implementation: preliminary model predictions are combined with climatological priors to sample possible states across weather to climate timescales, maintaining physical consistency throughout.}\label{fig:schematic}
\end{figure}

While often treated as separate tasks, assimilation and prediction share a fundamental connection --- both involve estimating the probability distribution of atmospheric states based on available information.
In assimilation (Fig.~\ref{fig:schematic}A left), we begin with the climatological distribution derived from historical observations. As new observations arrive, we continuously update this distribution, gradually narrowing down the range of possible states. 
In prediction (Fig.~\ref{fig:schematic}A right), this well-constrained distribution evolves back towards the climatological distribution through critical time ranges, until initial state information is blurred completely by chaos.

We implement this Probabilistic Assimilation and Prediction (PAP, Fig.~\ref{fig:schematic}A) framework through our Generative Assimilation and Prediction (GAP, Fig.~\ref{fig:schematic}B-D) methodology.
GAP chains
two key components:
1) a generative model that is trained on atmospheric reanalysis (Supplementary Information (SI) Sec. \textcolor{blue}{A}) to approximate the phase space distribution of atmospheric state (SI Sec. \textcolor{blue}{B.1});
and 2) a forecasting model that predicts the state evolution (SI Sec. \textcolor{blue}{B.2}). 
The generative component is a probabilistic diffusion model \cite{sohl2015deep,ho2020denoising,song2020score}, which learns target distributions by iteratively reversing a noise-adding process,
allowing flexible incorporation of observational, predictive, and external forcing constraints via approximate Bayesian computation \cite{song2020score, lugmayr2022repaint, zhang2023towards,meng2021sdedit,chao2024learning}. 
The forecasting component can be either a numerical or data-driven model -- we demonstrate GAP's capabilities with the latter, though either approach is viable.

In practice, GAP operates through a coherent cycle that seamlessly connects assimilation and prediction. At each assimilation step (Fig.~\ref{fig:schematic}B), observations constrain the generative model using a modified inpainting approach \cite{lugmayr2022repaint, zhang2023towards,chao2024learning}, 
which optimizes state estimates while ensuring consistency between observations and prior state information. 
These constrained estimates are then evolved forward by the forecasting model, creating preliminary predictions for the next time step. 
Rather than directly using these predictions, GAP applies calibrated noise and reverses the diffusion process \cite{meng2021sdedit, hess2024fast}, 
effectively introducing stochasticity that adheres to  the learned phase space structure, while simultaneously correcting systematic biases in the forecasting model (SI Sec. \textcolor{blue}{B.4}).

As we transition to prediction (Fig.~\ref{fig:schematic}C-D), this cycle continues without observational constraints, allowing the  well-constrained initial state distribution to gradually drift toward the climatological distribution. This evolution passes through distinct phases: weather forecasting (up to two weeks), where initial conditions dominate; seasonal-to-subseasonal prediction (up to one year), where ocean-atmosphere coupling provides extended predictability; and climate projection, where external forcings modulate the underlying climatological distribution. For these longer timescales, we leverage sea surface temperatures as constraints capturing both seasonal memory and climate change signals, effectively translating oceanic influences into appropriate atmospheric responses (SI Sec. \textcolor{blue}{B.5} and \textcolor{blue}{B.6}). This unified framework maintains probabilistic rigor and physical consistency across all prediction horizons without requiring separate training for different timescales.

\subsection*{Results}\label{Results}
We evaluate GAP's capabilities across five key aspects in weather-climate modeling:
(1) its ability to reproduce fundamental climatological patterns and variability, (2) its effectiveness in data assimilation for establishing accurate initial conditions, (3) its skill in weather forecasting, (4) its potential for seasonal prediction, and (5) its capacity to generate stable long-term climate simulations and projections.
Details on evaluation metrics and methods can be found in 
SI Sec. \textcolor{blue}{C}.



\subsection*{Climatology}\label{Climatology}

We first evaluate how the unconditional generation model in GAP reproduces the climatological distribution. This is  
similar to classical climate model evaluation.
Drawing from established climate model assessment frameworks \cite{klein2013climate,zhang2022e3sm,lee2024systematic}, 
we carry out a multi-scale comparison between  GAP's unconditional generation outputs with referential reanalysis data from European Center for Medium-Range Weather Forecasts (ECMWF) reanalysis v5 (ERA5) \cite{hersbach2020era5}.

\begin{figure}[H]
\centerline{\includegraphics[width=\linewidth]{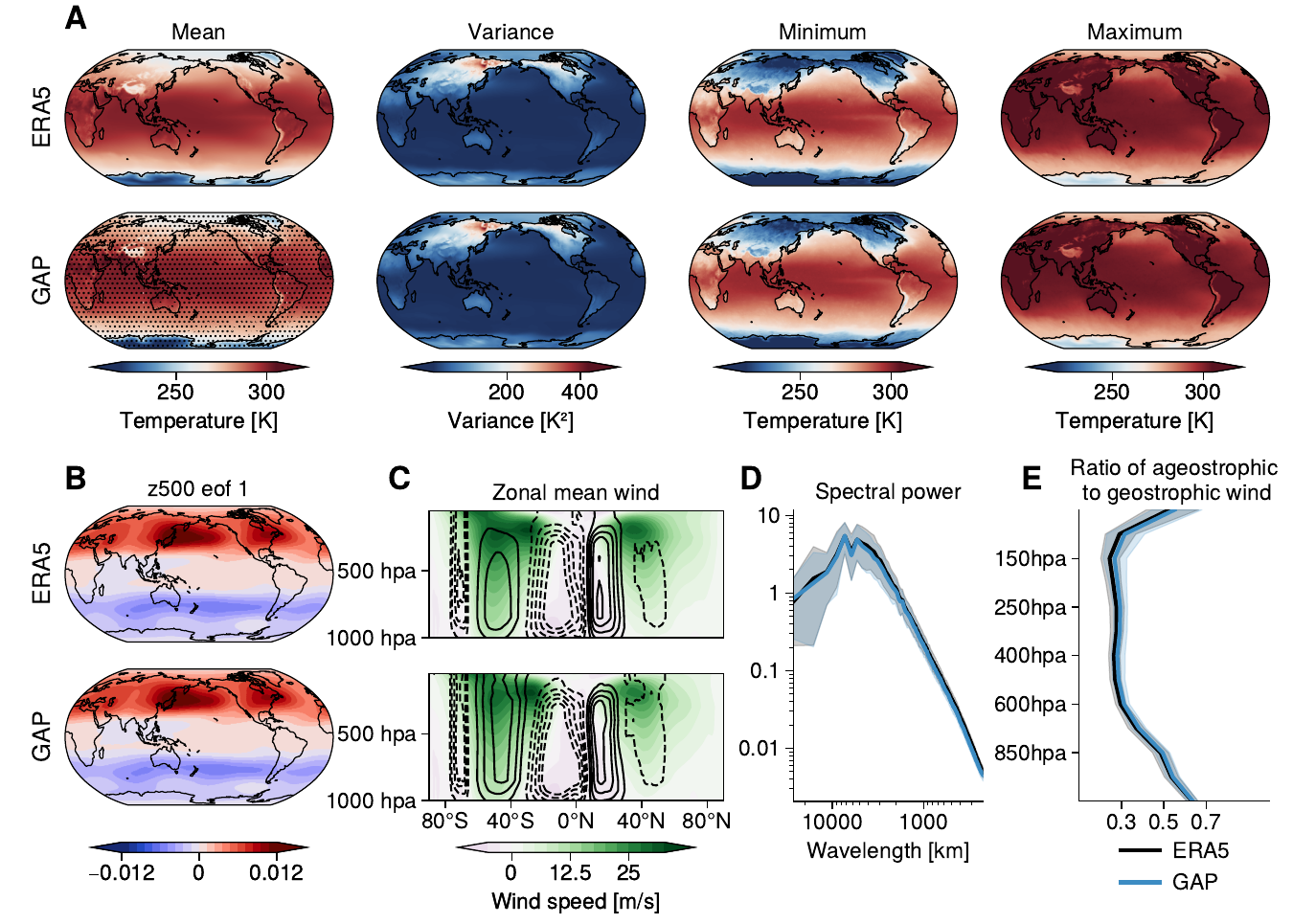}}
\caption{Multi-scale evaluation of GAP's climatological representation. \textbf{A}, Grid-point scale comparison between ERA5 and GAP using 2-meter temperature statistics (mean, variance, minimum, and maximum). Stippled areas indicate regions where GAP's generated distributions are statistically indistinguishable from ERA5 at the 95\% confidence level according to the Kolmogorov-Smirnov test. \textbf{B}, Leading empirical orthogonal function (EOF) of 500hPa geopotential height, revealing GAP's accurate reproduction of the Northern Annular Mode pattern. \textbf{C}, Hadley cell circulation represented by mass streamfunction (contours) and zonal-mean winds (colors) as a function of pressure and latitude. Solid (dashed) contours indicate positive (negative) values of the stream function, with contour intervals set to $2 \times 10^{10} \, \text{kg/s}$.  \textbf{D}, Spatial power spectral density as function of wavelength for 2-meter temperature, showing the characteristic power law decay from global to local scales. \textbf{E}, Vertical profiles of ageostrophic-to-geostrophic wind ratio in the extra-tropics ($|\text{lat}| \geq 20^\circ$), demonstrating preservation of fundamental atmospheric balance conditions.}\label{fig2}
\end{figure}

At the grid-point scale, GAP demonstrates exceptional fidelity in reproducing atmospheric statistical properties. 
Fig.~\ref{fig2}A presents a detailed comparison of key statistical measures (mean, variance, minimum, and maximum) for 2-meter temperature between ERA5 reanalysis (top row) and GAP's unconditional generation (bottom row). 
The model accurately captures spatial distribution patterns, including ocean-land temperature contrasts and topographical imacts, with particular precision in reproducing temperature extremes and variability. The stippled areas in Fig.~\ref{fig2}A areas of statistical agreement at 95\% confidence (SI~Sec. \textcolor{blue}{C.4}), which encompass most of the globe, particularly in regions characterized by complex temperature gradients.

This fidelity extends beyond temperature to the full suite of atmospheric variables. Analysis of spatially averaged relative differences (SI Fig. \textcolor{blue}{5}) confirms that GAP's reproduction of statistical measures remains remarkably accurate across all variables and pressure levels, with typical deviations below 10\% for most metrics. Comprehensive spatial patterns for additional variables are provided in SI Fig. \textcolor{blue}{6}-\textcolor{blue}{10}, showing consistently strong performance.

Beyond grid-point statistics, we evaluate GAP's ability to capture the organized structure of atmospheric dynamics across multiple scales. 
Analysis of the leading empirical orthogonal function of 500hPa geopotential height (Fig.~\ref{fig2}B) shows GAP precisely reproduces the Northern Annular Mode -- a dominant pattern that modulates jet stream position and storm tracks. This hemispheric-scale accuracy extends vertically, as demonstrated by GAP's faithful reproduction of the Hadley circulation (Fig.~\ref{fig2}C), capturing both the tropical cell's strength and vertical extent, which drives global energy redistribution.

Across spatial scales, GAP maintains consistent fidelity. The spherical harmonic power spectra analysis (Fig.~\ref{fig2}D) confirms that GAP correctly reproduces the characteristic power law decay from global to local scales, preserving both large-scale patterns and their cascading influence on smaller scales (SI Fig. \textcolor{blue}{11}-\textcolor{blue}{12} provides results for additional variables). 
Finally, GAP maintains proper physical relationships between variables, as shown by the nearly identical vertical profiles of ageostrophic-to-geostrophic wind ratios in the extra-tropics (Fig.~\ref{fig2}E), indicating preservation of fundamental atmospheric balance conditions.

This successful approximation of Earth's climatological distribution provides the foundation for GAP's capabilities in data assimilation and seamless prediction across timescales.




\subsection*{Assimilation}\label{Assimilation}

Data assimilation aims to combine prior predictive information with new-coming observations to obtain optimal state estimates, 
so as to initialize a forecast \cite{evensen2022data}.
While conventional methods rely on various approximations to make this process computationally tractable \cite{fisher2001developments, houtekamer2016review}, GAP directly addresses the probabilistic nature of this challenge by updating state estimates given predictive prior and observational evidence in a sequential Bayesian inference manner (see Methods).

To validate our approach, we conduct two parallel experiments: GAP-OBS, which uses real observations from global radiosondes and surface stations (data details given in SI Sec. \textcolor{blue}{A.3}), and GAP-OSSE (Observing System Simulation Experiments) \cite{arnold1986observing}, which uses ``perfect'' observations from reanalysis data, so as to isolate observational uncertainty effects. We use initial conditions from ECMWF's Integrated Forecast System High-Resolution Forecast (IFS-HRES) \cite{buizza2018development} --- the leading operational assimilation and prediction system --- as our benchmark.

\begin{figure}[H]
\centering
\centerline{\includegraphics[width=\linewidth]{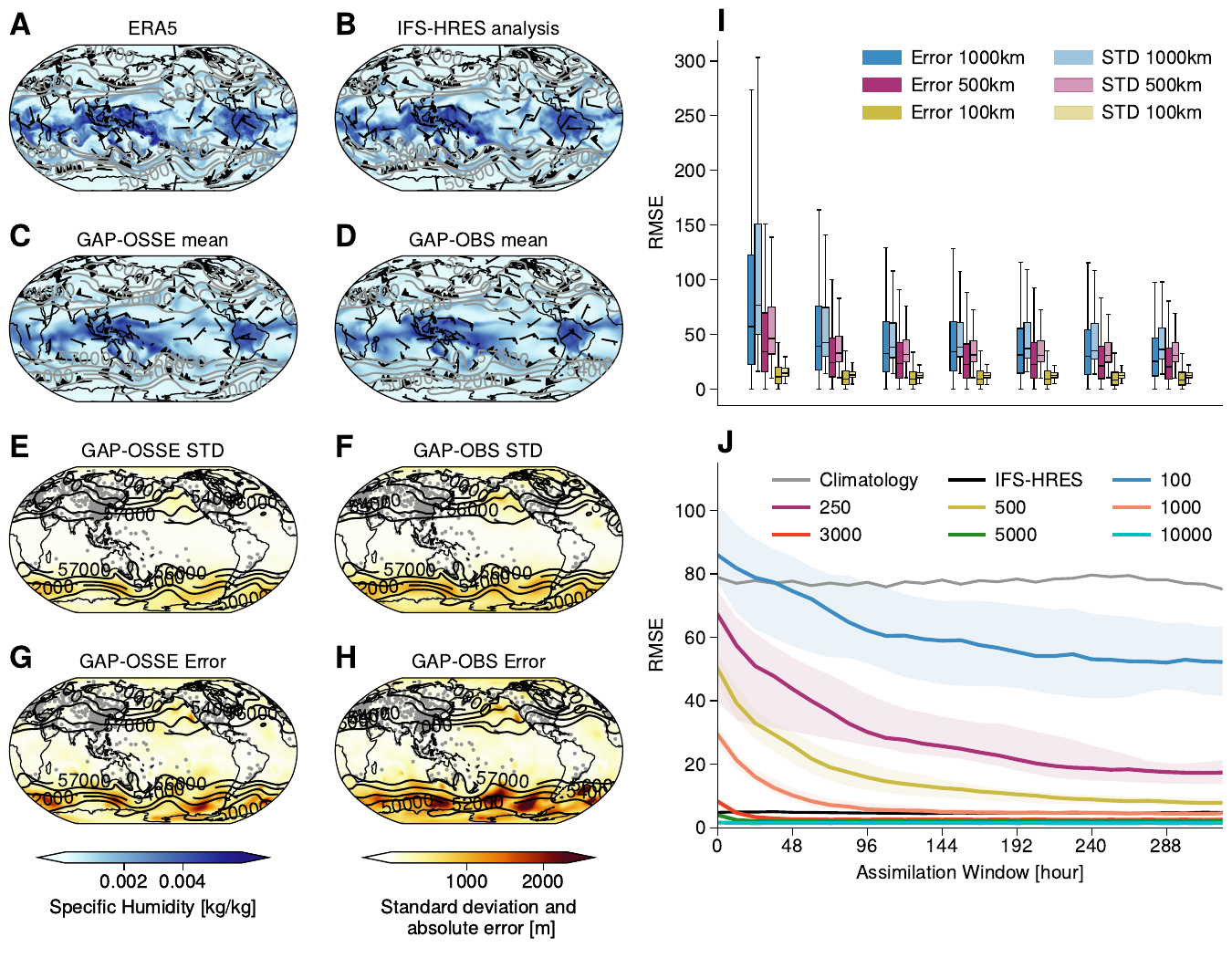}}
\caption{Evaluation of GAP's data assimilation capabilities. \textbf{A-D}, Spatial patterns of specific humidity and geopotential height from ERA5 (ground truth), IFS-HRES analysis (operational benchmark), GAP-OSSE (perfect observations), and GAP-OBS (real observations), demonstrating comparable accuracy between GAP and operational systems. \textbf{E-H}, Ensemble standard deviation and error patterns from GAP-OSSE and GAP-OBS, showing strong correspondence between predicted uncertainty (spread) and actual error, indicating reliable uncertainty quantification. \textbf{I}, RMSE statistics as a function of distance from observation points through the assimilation window, illustrating how error (deep colors) and ensemble variance (shallow colors) decrease near observations and increase with distance. \textbf{J}, RMSE skill at different observation density levels through the assimilation window for GAP ensemble (blue line with shading) and IFS-HRES analysis (black line), showing progressive acceleration in convergence as observation density increases from climatological levels to $10^4$ observational points.}\label{fig_assimilation}
\end{figure}

Detailed analysis of both experiments reveals GAP's remarkable capabilities in state estimation and uncertainty quantification. Fig.~\ref{fig_assimilation}A-D show how the ensemble mean fields in both GAP-OBS and GAP-OSSE experiments converge to closely match the reanalysis state, achieving accuracy comparable to IFS-HRES initial conditions (see SI Fig. \textcolor{blue}{13} for full trajectories through the assimilation window). A key strength of GAP emerges in its uncertainty representation: the ensemble spread patterns (Fig.~\ref{fig_assimilation}E-F) show very good correspondence with actual error in the ensemble mean (Fig.~\ref{fig_assimilation}G-H), both in spatial distribution and magnitude. This alignment between predicted uncertainty and actual error -- known as the spread-skill relationship \cite{scherrer2004analysis} -- indicates GAP's ability to provide reliable uncertainty estimates \cite{nai2024reliable}, a crucial capability that has proven challenging for conventional methods \cite{evensen2022data}. A comparison between the GAP-OBS and GAP-OSSE experiments reveals the impact of observational quality. While both achieve similar final accuracy, GAP-OSSE exhibits slightly tighter spread and error patterns due to its ``perfect'' observations, while GAP-OBS demonstrates robustness to real-world observational uncertainties without sacrificing reliability.

As new observations arrive, GAP dynamically updates its ensemble of possible atmospheric states. Fig.~\ref{fig_assimilation}I illustrates a fundamental pattern: ensemble variance naturally decreases around observation points while growing with distance. Ensemble mean error follow a similar decreasing rate (see SI Fig. \textcolor{blue}{14} for a comprehensive assessment). 
This spatial structure of uncertainty provides a physically meaningful measure of confidence in the analysis - lower near observations and higher in data-sparse regions - reflecting how information from observations diffuses through space, and reflecting the capability of GAP to reliably capture the associated uncertainties.

A systematic evaluation across varying observational densities reveals an intricate interplay between observational constraints and chaos-induced uncertainty growth \cite{murphy2024duality}, which is revealed in Fig.~\ref{fig_assimilation}J (see also SI Fig. \textcolor{blue}{15}): as observation density increases from 0 (climatological level) to $10^4$ observational points, 
we observe progressively faster convergence to IFS-HRES accuracy (black line). This pattern demonstrates how denser observations more effectively constrain the system's inherent chaotic tendency to diverge. At lower densities, multiple assimilation cycles are needed to accumulate sufficient observational constraints, while very dense observations can immediately constrain the state, essentially overcoming chaos through direct measurement.


A year-long assessment across multiple atmospheric variables for year 2020 (Fig.~\ref{fig_metrics}A) confirms that GAP achieves assimilation accuracy comparable to IFS-HRES analysis using less than 1\% of typical operational observations. These results establish GAP as a significant advance in data assimilation, achieving state-of-the-art accuracy with remarkable efficiency while maintaining reliable uncertainty estimates across diverse observational conditions.

\subsection*{Weather forecast}\label{Weather_forecast}

\begin{figure}[H]
\centering
\centerline{\includegraphics[width=.95\linewidth]{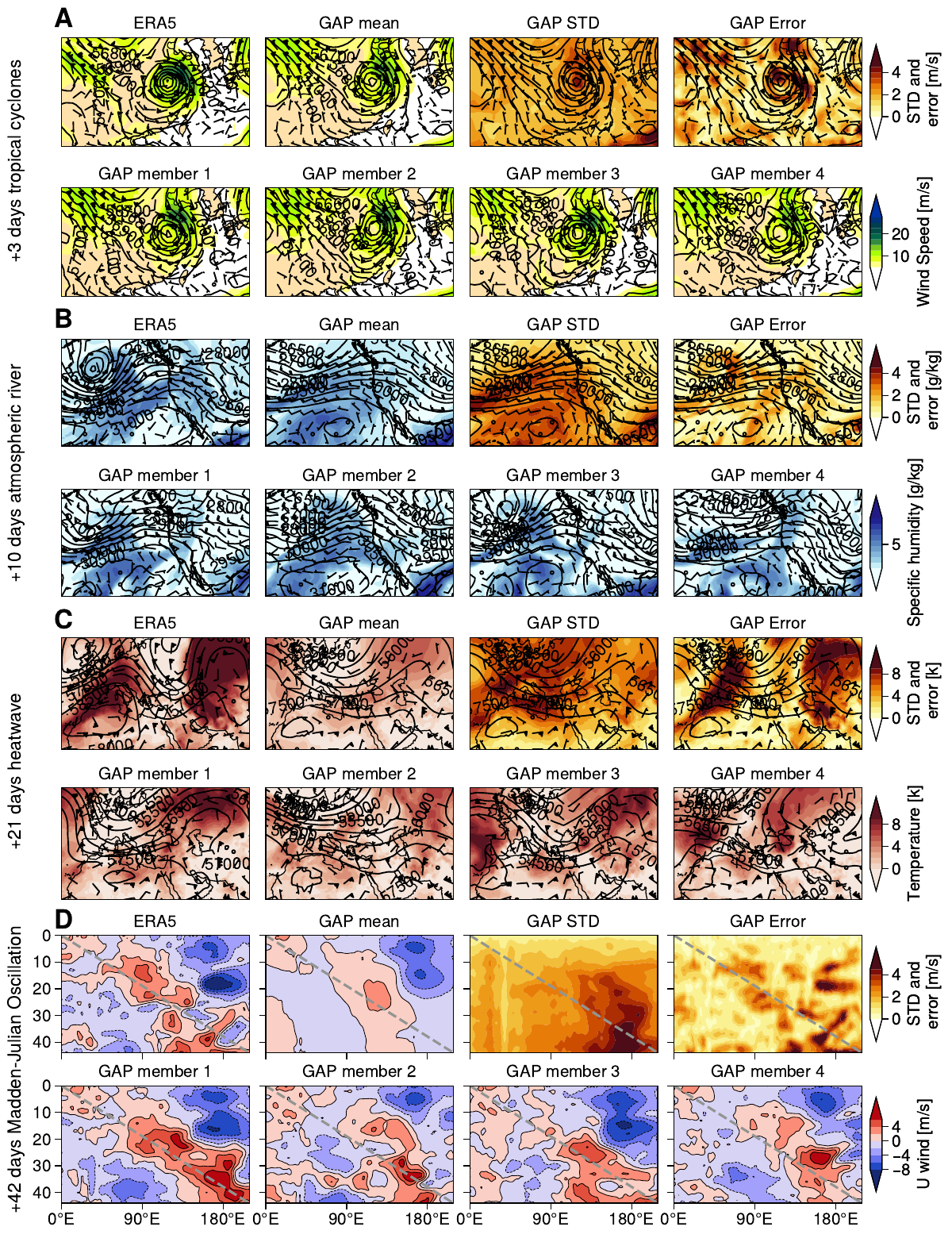}}
\caption{GAP's ensemble forecasts of high-impact weather events.
\textbf{A}, 3-day forecast of typhoon In-Fa (July 28, 2021), showing wind speed (color shading) and geopotential height (contours) at 500hpa from ERA5 reference, GAP ensemble mean, standard deviation, ensemble mean prediction error, and individual ensemble members. 
\textbf{B}, 10-day forecast of an atmospheric river hitting California (April 12, 2023), displaying 850hPa geopotential height (contours) and specific humidity (color shading). 
\textbf{C}, 21-day forecast of a European heatwave (June 23, 2023), showing 2-meter temperature anomalies (color shading) and 500hPa geopotential height (contours).
\textbf{D}, 42-day forecast of the Madden-Julian Oscillation (January 20 to March 1, 2023), presenting Hovm\"{o}ller diagram of zonal wind anomalies at 850hPa (color shading) across tropical latitudes (0-45$^\circ$S) and longitudes (0-180$^\circ$E).}\label{weather_forecast_case}
\end{figure}

\begin{figure}[H]
\centering
\centerline{\includegraphics[width=\linewidth]{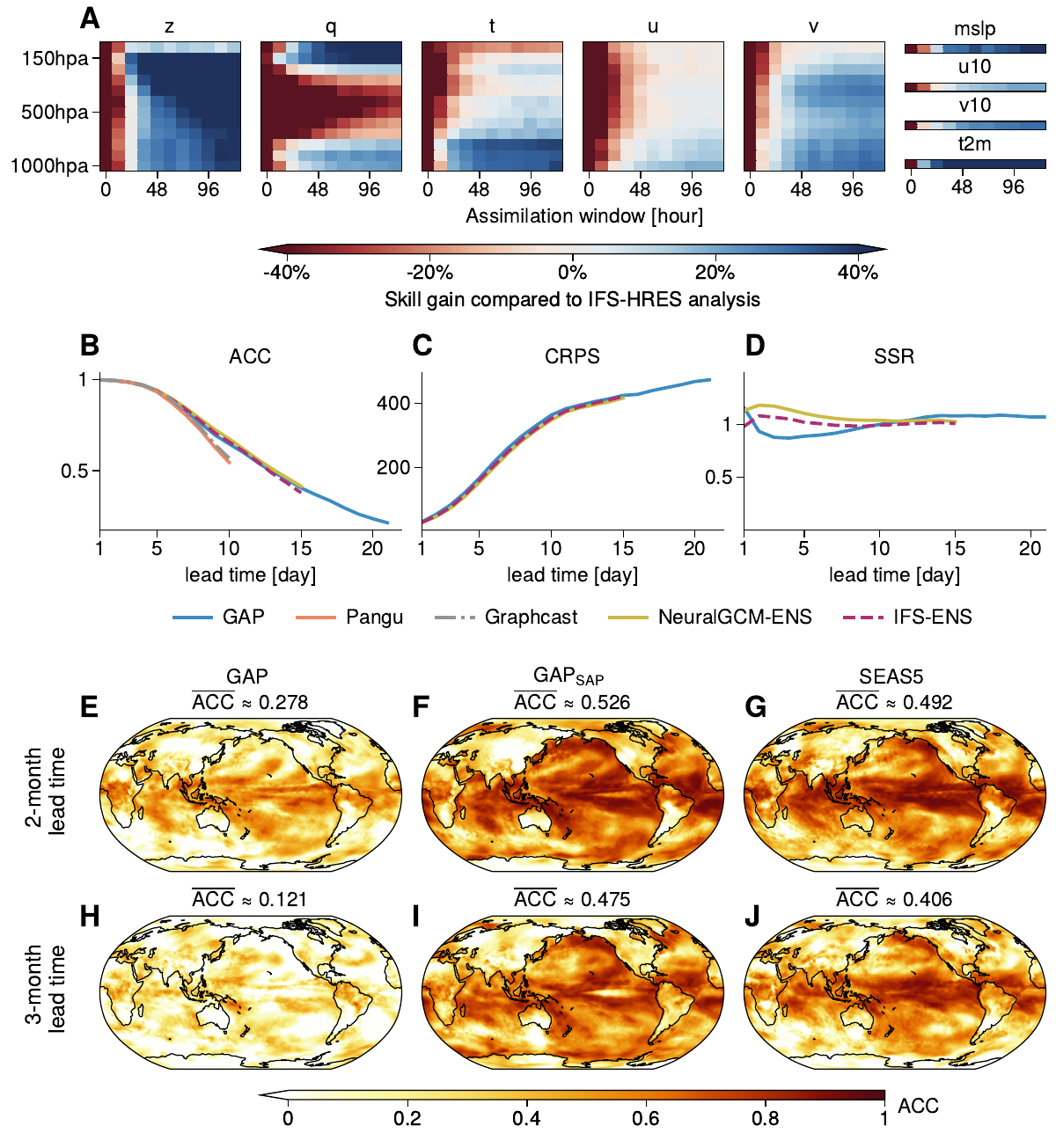}}
\caption{Evaluation of GAP's performance across assimilation and prediction tasks.
\textbf{A}, Comparing GAP assimilation versus IFS-HRES initial conditions across multiple variables, pressure levels, and assimilation times, with blue/red cells indicating reduced/increased assimilation error compared to IFS-HRES analysis (measured by relative RMSE difference, see SI Sec. \textcolor{blue}{C.1}). 
\textbf{B-D}, Quantitative comparison of medium-range forecast skill between GAP, IFS-ENS, NeuralGCM-ENS, Pangu, and GraphCast, showing: \textbf{B}, anomaly correlation coefficient (deterministic skill); \textbf{C}, continuous ranked probability score (probabilistic skill); and \textbf{D}, spread-skill ratio (uncertainty calibration). 
\textbf{E-J}, Spatial distribution of seasonal prediction skill for 2-meter temperature at 2-month (\textbf{E-G}) and 3-month (\textbf{H-J}) lead times, comparing GAP without SST forcing (\textbf{E},\textbf{H}), $\text{GAP}_{\text{SAP}}$ with SST anomaly persistence (\textbf{F},\textbf{I}), and ECMWF's SEAS5 (\textbf{G},\textbf{J}). Latitude-weighted mean ACC values are indicated at the top of each panel.}
\label{fig_metrics}
\end{figure}

We evaluate GAP's weather forecasting capability via two complementary challenges: probabilistic prediction of high-impact weather events (Fig.~\ref{weather_forecast_case}), and systematic comparison with cutting-edge numerical weather prediction \cite{buizza2018development}, ML \cite{bi2023accurate,lam2023learning}, and hybrid \cite{kochkov2024neural} forecasting systems (Fig.~\ref{fig_metrics}B-D).



We consider forecasts for high-impact weather events, including
3-day-ahead prediction of typhoons (Fig.~\ref{weather_forecast_case}A), 
10-day-ahead prediction of atmospheric rivers (Fig.~\ref{weather_forecast_case}B), 
21-day-ahead prediction of heat waves (Fig.~\ref{weather_forecast_case}C), 
and 
42-day-ahead prediction of the Madden Julian Oscillation (MJO, Fig.~\ref{weather_forecast_case}D). 
We compare GAP's ensemble predictions with reanalysis, presenting ensemble means, standard deviations, ensemble mean prediction errors, and randomly selected ensemble members (see SI Fig. \textcolor{blue}{17}-\textcolor{blue}{20} for additional ensemble members). 
For the case of Typhoon In-Fa (Fig.~\ref{weather_forecast_case}A), a severely devastating tropical cyclone that occurred in July 2021 over the Northwest Pacific Ocean,
GAP accurately captures both the cyclone's structural evolution through wind and pressure fields, and its position uncertainty through ensemble spread. While the ensemble mean appears diffuse, individual ensemble members show sharp, physically realistic structures that closely align with reanalysis, demonstrating GAP's ability to generate credible probabilistic forecasts without explicit dynamical constraints. 
The atmospheric river forecast (Fig.~\ref{weather_forecast_case}B) extends to a 10-day horizon, where GAP maintains physical consistency in predicting both 850 hPa geopotential height patterns and specific humidity distributions -- a capability that distinguishes it from pure pattern-matching approaches \cite{bi2023accurate,lam2023learning} and illustrates its grasp of moisture transport processes and mid-latitude dynamics. The European heatwave prediction (Fig.~\ref{weather_forecast_case}C) pushes beyond conventional forecasting limits, maintaining skill through day 21 in capturing both temperature anomalies and associated circulation patterns. This extended-range capability indicates that GAP has learned to identify and preserve slow-evolving atmospheric patterns that enhance predictability. Finally, the 42-day MJO forecast (Fig.~\ref{weather_forecast_case}D) reveals GAP's ability to predict large-scale organized weather events, establishing a seamless forecast that accurately represents both the eastward propagation and amplitude modulation of this tropical phenomenon. 

We conduct a year-long  evaluation for year 2020 (Fig.~\ref{fig_metrics}B-D),  comparing GAP against state-of-the-art weather prediction systems spanning three distinct approaches: physics-based operational ensemble forecasting (IFS-ENS \cite{buizza2018development}), pure ML models (Pangu \cite{bi2023accurate} and GraphCast \cite{lam2023learning}), and a hybrid system combining a process-based dynamical core with ML parameterization (NeuralGCM \cite{kochkov2024neural}). For this comparison, we focus on the 500 hPa geopotential height (z500) --- a key atmospheric field that characterizes mid-tropospheric weather patterns and serves as a standard benchmark in weather prediction (see SI Fig. \textcolor{blue}{21}-\textcolor{blue}{23} for extra evaluation results).

GAP demonstrates remarkable advantages across the considered deterministic and probabilistic metrics with notable computation efficiency (SI Sec. \textcolor{blue}{B.1.5}). Regarding deterministic skill, measured by the Anomaly Correlation Coefficient (ACC, Fig.~\ref{fig_metrics}B), GAP outperforms pure ML approaches throughout the forecast period, showing 3-14\% improvement in forecast skill. This advantage is particularly notable beyond day 7, where the accuracy of pure ML models deteriorates rapidly. The Continuous Ranked Probability Score (CRPS) analysis (Fig.~\ref{fig_metrics}C) reveals that GAP achieves probabilistic forecast skill comparable to the world-leading IFS-ENS system \cite{buizza2018development}, as well as the hybrid NeuralGCM approach \cite{kochkov2024neural}, particularly in the critical 5-15 day range where weather prediction uncertainty grows significantly. The Spread-Skill Ratio (SSR) analysis (Fig.~\ref{fig_metrics}D) provides crucial insight into GAP's uncertainty quantification capabilities. While IFS-ENS \cite{buizza2018development} maintains a nearly optimal spread-skill ratio (close to 1) through careful ensemble design, GAP achieves similar reliability through its generative approach. Although GAP's spread is wider in the first few days, it gradually converges to optimal calibration, suggesting the model has learned the natural growth rates of forecast uncertainty. This performance is particularly noteworthy because GAP achieves this calibrated uncertainty representation without the extensive tuning required in traditional ensemble systems.

These results demonstrate that GAP's generative framework successfully combines the computational efficiency of ML approaches  with the physical consistency and uncertainty awareness of numerical ensemble systems, while overcoming the scale separation issues inherent in hybrid models \cite{kochkov2024neural}.

\subsection*{Potential for seasonal prediction}\label{seasonal_forecast}
Here we focus on seasonal prediction, during which initial atmospheric state information gets diminished by chaos, but slowly evolving climate system components, particularly sea surface temperature (SST), serve as sources of predictability through their persistent thermal memory and ability to constrain atmospheric evolution.

We introduce $\text{GAP}_{\text{SAP}}$ (GAP constrained by SST Anomaly Persistence), which incorporates SST forcing through a straightforward approach of persisting initial SST anomalies throughout the forecasting period (SI Sec.  \textcolor{blue}{B.5}). While simple, this design allows us to assess GAP's capability to translate oceanic conditions into atmospheric responses.
We conduct a 30-year re-forecast experiment comparing three configurations: the original GAP (without SST forcing), $\text{GAP}_{\text{SAP}}$, and ECMWF's latest, fifth generation of  its seasonal forecasting system SEAS5. 


The comparison between GAP (Fig.~\ref{fig_metrics}E and H)  and $\text{GAP}_{\text{SAP}}$ (Fig.~\ref{fig_metrics}F and I) demonstrates the fundamental value of SST forcing in seasonal prediction (see SI Fig.  \textcolor{blue}{24}-\textcolor{blue}{28} for comprehensive comparison). $\text{GAP}_{\text{SAP}}$ achieves notably higher ACCs than the baseline GAP at both 2-month and 3-month lead times ($\overline{\text{ACC}}\approx0.526$ v.s. 0.278 at 2 months; 0.475 v.s. 0.121 at 3 months). This dramatic improvement indicates that GAP's generative framework can effectively utilize boundary condition information to enhance predictive skill. The spatial distribution of this improvement is particularly informative – while skill increases are observed globally, they are most pronounced in regions with strong ocean-atmosphere coupling, suggesting a successful capturing of key air-sea interaction processes.

When comparing $\text{GAP}_{\text{SAP}}$ (Fig.~\ref{fig_metrics}F and I) with SEAS5 (Fig.~\ref{fig_metrics}G and J), we observe a more nuanced picture that reveals both the strengths and limitations of the current approach. At a global level, $\text{GAP}_{\text{SAP}}$ surpasses SEAS5's performance by a large margin ($\overline{\text{ACC}}\approx0.526$ v.s. 0.492 at 2 months; 0.475 v.s. 0.406 at 3 months), but with distinct regional variations (SI Fig. \textcolor{blue}{28}). Over land regions, such as Eurasia, $\text{GAP}_{\text{SAP}}$ shows poorer skill in capturing year-to-year temperature variations. In ocean regions, such as the North Pacific, $\text{GAP}_{\text{SAP}}$ frequently exceeds SEAS5's performance, indicating particularly effective translation of oceanic conditions into atmospheric responses.

The spatial pattern of seasonal prediction skill illuminates key sources of uncertainty in $\text{GAP}_{\text{SAP}}$'s methodology, which points out directions for further improvement. First, the simple SST anomaly persistence approach introduces errors in predicting SSTs (SI Sec.  \textcolor{blue}{B.5.2}), particularly in dynamically active regions like the tropical Pacific, where the dominant signal of ocean dynamic variability, i.e., the El Ni$\mathrm{\tilde{n}}$o-Southern Oscillation (ENSO), is poorly captured by persistence \cite{zhao2021correspondence}. Second, the process of inferring sea air temperatures from SST adds another layer of uncertainty (SI Sec. \textcolor{blue}{B.5.5}). Third, the inpainting-based atmospheric constraint may not optimally propagate information from ocean to land regions. These compound uncertainties likely explain why $\text{GAP}_{\text{SAP}}$'s skill, while substantially better than GAP's, shows more pronounced degradation over continental interiors compared to SEAS5's fully coupled approach.

These results suggest that, even with a simple implementation of SST forcing, our framework can capture essential aspects of seasonal predictability. The success of $\text{GAP}_{\text{SAP}}$, particularly in ocean-dominated regions, indicates significant potential for further improvements through more sophisticated approaches to both dynamical ocean prediction and ocean-atmosphere coupling.

\subsection*{Climate simulation}\label{climate_simulation}


Finally, we investigate GAP's performance in climate simulation and projection, which requires three critical capabilities that challenge both physics-based and machine learning approaches: numerical stability over extended timescales, accurate representation of climate variability across temporal scales, and appropriate response to external forcings. Our assessment framework evaluates these capabilities systematically, revealing GAP's significant advances in climate modeling.

\begin{figure}[H]
\centering
\centerline{\includegraphics[width=1.0\linewidth]{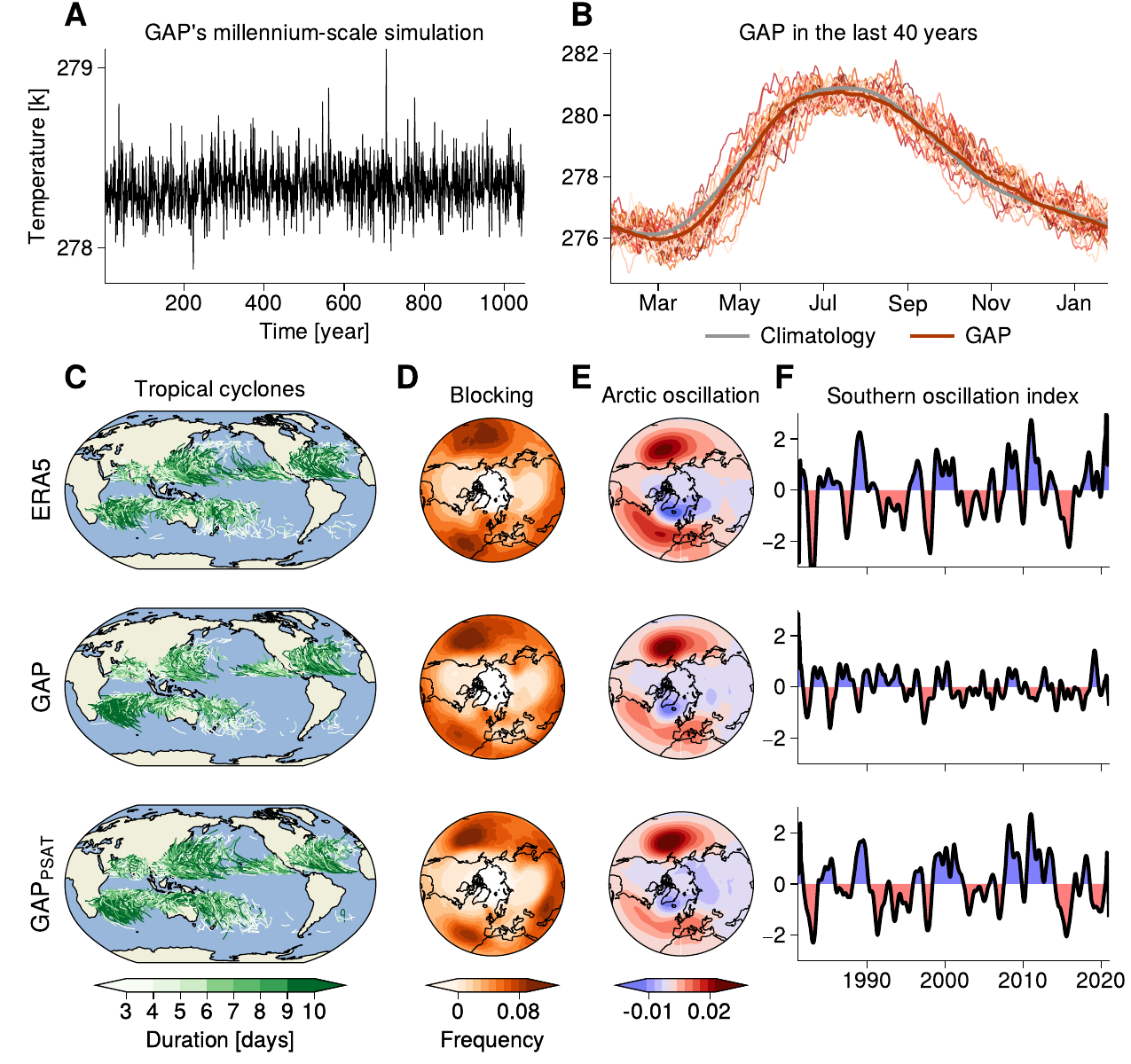}}
\caption{Evaluation of GAP's climate simulation capabilities. 
\textbf{A}, Millennial-scale simulation stability shown through global mean surface air temperature for a 1050-year free run (assimilated from 1980, extending to 3030). 
\textbf{B}, Seasonal cycle reproduction in the final 40 years (2990-3030), with individual years (thin red lines) and climatological average (thick red line) compared to ERA5 climatology (blue line). \textbf{C-F}, Comparison of climate variability features between ERA5 (top row), GAP free run (middle row), and GAP with prescribed sea-air temperatures ($\text{GAP}_\text{PSAT}$, bottom row): 
\textbf{C}, tropical cyclone track density with duration indicated by colors; 
\textbf{D}, mid-latitude blocking frequency; 
\textbf{E}, Arctic Oscillation pattern represented by sea level pressure regression; 
and \textbf{F}, Southern Oscillation Index time series demonstrating interannual variability.}
\label{fig_climate}
\end{figure}

Our millennium-scale free-running simulation demonstrates GAP's exceptional numerical stability—a persistent challenge for ML-based climate models. Without any external constraints, GAP maintains physically realistic temperature variability around a stable mean for over 1,000 years (Fig.~\ref{fig_climate}A), avoiding the drift that typically plagues long-term climate simulations. This stability extends to seasonal cycles, with GAP accurately reproducing the characteristic amplitude and phase of annual temperature variations even after a millennium of simulation (Fig.~\ref{fig_climate}B).

Beyond basic stability, GAP successfully captures the complex modes of climate variability. The model reproduces key atmospheric phenomena across multiple scales, including tropical cyclone spatial distributions and intensity characteristics (Fig.~\ref{fig_climate}C), mid-latitude blocking patterns that regulate weather persistence in the extratropics (Fig.~\ref{fig_climate}D), as well as the Arctic Oscillation (Fig.~\ref{fig_climate}E) -- a dominant mode of atmospheric variability affecting Northern Hemisphere weather patterns. 
The free-running model shows limitations in representing phenomena strongly tied to ocean forcing, such as the Southern Oscillation Index (Fig.~\ref{fig_climate}f). However, when provided with prescribed historical sea air temperatures ($\text{GAP}_\text{PSAT}$), GAP successfully reproduces these ocean-driven patterns, demonstrating appropriate atmospheric response to boundary conditions.


To evaluate GAP's response to climate change scenarios, we conduct ensemble simulations using sea surface temperatures from the community earth system model version 2 (CESM2) \cite{danabasoglu2020community} future scenario projections as boundary conditions, following three trajectories of Shared Socioeconomic Pathways (SSP1-2.6, SSP3-7.0, and SSP5-8.5). 
These experiments demonstrate that GAP maintains stable atmospheric structure and circulation patterns (SI Fig. \textcolor{blue}{33}-\textcolor{blue}{36}) while significantly reducing typical climate model biases (SI Fig. \textcolor{blue}{37}). Furthermore, GAP accurately reproduces extreme weather events and key modes of climate variability (SI Fig. \textcolor{blue}{38}-\textcolor{blue}{41}). Additionally, our analysis reveals that, while GAP effectively captures the expected shifts in temperature profiles under different climate scenarios, it nonetheless exhibits some generalization challenges common to other ML models (SI Fig. \textcolor{blue}{42} and \textcolor{blue}{43}).

These results establish GAP as a powerful framework for climate simulation, achieving stable millennial-scale integrations, accurately reproducing climate variability across temporal scales, and appropriately responding to external forcings under different scenarios. By combining these capabilities with the computational efficiency of machine learning methods, GAP offers new opportunities for rapid climate simulations, comprehensive scenario exploration, and ensemble-based climate projections.

\section*{Discussion}\label{discussion}
The emergence of machine learning in weather forecasting has demonstrated unprecedented computational efficiency while maintaining accuracy for short-term predictions. However, the fundamental pursuit of climate science extends beyond sequential state prediction --- it seeks to understand how Earth's climate evolves as a chaotic, high-dimensional, multi-scale system. This understanding is crucial for monitoring atmospheric conditions and predicting their evolution across a broad range of spatiotemporal scales, from local convection to global circulation patterns, from weather variability to climate change responses.

GAP marks an advance in weather and climate modeling by unifying data assimilation, seamless forecasting, and climate simulation within a generative framework. By recognizing the duality between assimilation and prediction \cite{murphy2024duality,evensen2022data}, while leveraging the asymptotic distribution of chaotic systems as a powerful pattern recognizer \cite{gilpin2024generative,chao2024learning,pan2021learning,pan2022improving}, GAP leverages machine learning efficiency to tackle fundamental challenges in weather and climate science. 
This yields practical advances across five critical areas: (1) reproduction of fundamental climatological patterns and variability, (2) effective and efficient data assimilation for establishing accurate initial conditions, (3) skillful weather forecasting, (4) promising seasonal prediction, and (5) stable long-term climate simulations and projections in response to forcing.

When compared with current approaches, GAP offers distinct advantages. Traditional physics-based models continue evolving through enhanced resolution and improved parameterization schemes, yet their development cycle is increasingly constrained by error compensation and diagnostic complexity \cite{stevens2013climate,palmer2016personal,stevens2024perspective}. Pure machine learning approaches, while efficient,  mostly focus on short-term prediction tasks, but do not incorporate
the necessary data assimilation. Moreover, these models tend to underestimate extreme weather events, and suffer from error accumulation in long roll-outs \cite{lam2023learning, price2024probabilistic, rackow2024robustness, hakim2024dynamical}. Hybrid approaches like NeuralGCM \cite{kochkov2024neural}, which combine dynamical cores with neural network parameterizations, face significant scalability challenges and struggle to explicitly model the response to external forcing \cite{kochkov2024neural}.
GAP's architecture offers a more versatile solution: its generative framework can accommodate either dynamical cores for physical consistency, or data-driven forecasting for computational efficiency, while maintaining its core capability to assimilate various observations, model response to external forcing, and scale efficiently. While our current implementation demonstrates the advantages of the data-driven approach, integrating dynamical cores within this flexible framework remains an active area of our research. Additional important directions include expanding beyond state-corresponding observations to more complex remote-sensing data, and extending our treatment of forcings beyond prescribed sea surface temperature to include ocean dynamics, as well as greenhouse gas and aerosol forcings.



GAP's success also reveals profound connections between generative modeling and chaotic dynamics: both essentially transform random inputs into structured patterns. The climate system, with its well-documented chaotic behavior and rigorous prediction metrics, provides a more demanding test for generative models than traditional benchmarks \cite{gilpin2021chaos}. This suggests exciting possibilities for cross-pollination between chaos theory, climate science, and machine learning.

\newpage

\section*{Methods}\label{Methods}













\subsection*{Probabilistic assimilation and prediction}\label{Problem_statement}
We tackle the task of assimilation and prediction in a unified formulation, i.e., estimating the probabilistic distribution of atmospheric states given observational, predictive, and external forcing constraints.

Let $\mathbf{X}\in \mathbb{R}^d$ represent the atmospheric state vector, which comprises five primitive variables (zonal and meridional wind velocity, temperature, specific humidity, and geopotential height) defined on a three-dimensional grid (latitude$\times$longitude$\times$pressure levels), along with four surface variables (10-meter winds, 2-meter temperature,  and sea-level pressure) defined over latitude and longitude spans (SI Table \textcolor{blue}{1}). 
At our current resolution of $1^\circ\times1^\circ$ with 13 pressure levels, the state space dimension 
$d\approx 4.5$ million\footnote{$d=(\underbrace{5}_{\text{upper variable}}\times\underbrace{13}_{\text{pressure level}}+\underbrace{4}_{\text{surface variable}})\times\underbrace{181}_{\text{latitude}}\times\underbrace{360}_{\text{longitude}} =4496040$}.

We now define three key timestamps that delineate different phases in an assimilation and prediction cycle:
\begin{itemize}
    \item $t_a$: The start of the assimilation window, when observations begin to constrain state uncertainty.
    \item $t_0$: 
    The transition point between assimilation and prediction, after which no new observations are available.
    \item $t_p$: 
    The end of the prediction window, which may extend to climate timescales.
\end{itemize}
Given this setup, we address the following three interconnected estimation tasks:

\begin{enumerate}
    \item \textbf{Climatological distribution estimation}:
    For $t<t_a$, we estimate the baseline climatological distribution by maximizing 
    \begin{equation*}
        p_{\theta^{*}}=\argmax_{p_\theta}{\prod_{i=1}^{N} p_\theta(\mathbf{X}_i|\Theta)},
    \end{equation*}
    where  
    $p_{\theta}$ is a parameterized probability distribution with parameters $\theta$,
    $\mathbf{X}_{i\in[1,N]}$ are atmospheric state samples from climate reanalysis, 
    and $\Theta$ represents external forcing conditions. 
    \item \textbf{Assimilation}: For $t=t_a$, given observations $\mathbf{O}_{t_a}$ and an observation operator defining $p(\mathbf{O}|\mathbf{X})$, we estimate 
    the conditional distribution
    \begin{equation*}
        p(\mathbf{X}_{t_a}|\mathbf{O}_{t_a})\propto 
        p_{\theta^*}(\mathbf{X}|\Theta)p(\mathbf{O}_{t_{a}}|\mathbf{X}_{t_a})
    \end{equation*}
    For $t\in[t_a+\Delta t,t_0]$, where $\Delta t$ denotes the computation time step, 
    given a sequence of observations $\mathbf{O}_{t_a:t}$, and a forecasting model $\mathbf{M}$ that induces transition $p(\mathbf{X}_{t+\Delta t}|\mathbf{X}_{t})$,
    we then estimate the evolving state distribution through recursive Bayesian updates:

    \begin{equation*}
        p(\mathbf{X}_{t}|\mathbf{O}_{t_a:t})=
        \int p(\mathbf{X}_{t}|\mathbf{X}_{t-\Delta t},\mathbf{O}_{t})
        p(\mathbf{X}_{t-\Delta t}|\mathbf{O}_{t_a:{t-\Delta t}})d\mathbf{X}_{t-\Delta t}
    \end{equation*}

    \item \textbf{Prediction}: For $t>t_0$, starting from the assimilated distribution $p(\mathbf{X}_{t_{0}}|\mathbf{O}_{t_a:t_0})$, project forward to estimate:
    \begin{equation*}
    p(\mathbf{X}_{t_{0:p}}|\mathbf{X}_{t_{0}},\Theta)=\prod_{t=t_0}^{t_p-\Delta t}
        p(\mathbf{X}_{t+\Delta t}|\mathbf{X}_{t},\Theta)
    \end{equation*}
    This prediction spans multiple characteristic ranges from weather to climate, with dominant source of predictability shifting from the initial state $\mathbf{X}_{t_0}$ to the external forcing $\Theta$. 
\end{enumerate}


These three tasks are  linked through their dependence on the underlying climatological distribution, with each task addressing distinct sources of uncertainty — from observational uncertainties in data assimilation, to chaos-induced growth of forecast uncertainty in prediction, to forcing-dependent variability in long-term climate projections. Together, they constitute the Probabilistic Assimilation and Prediction (PAP) framework.

\subsection*{Generative Assimilation and Prediction}\label{GAP_methods}
GAP offers practical solution to the PAP formulation 
regarding the computation of the distributions $p_{\theta^*}(\mathbf{X}|\Theta)$, $p(\mathbf{X}_{t_a}|\mathbf{O}_{t_a})$, $p(\mathbf{X}_{t}|\mathbf{X}_{t-\Delta t},\mathbf{O}_{t})$, $p(\mathbf{X}_{t}|\mathbf{X}_{t-\Delta t})$, and $p(\mathbf{X}_{t}|\mathbf{X}_{t-\Delta t},\Theta)$, as explained in detail below.

\subsubsection*{$p_{\theta^*}(\mathbf{X}|\Theta)$: climatological distribution estimation}
To approximate the climatological distribution of atmospheric states, we parameterize $p_{\theta^*}(\mathbf{X}|\Theta)$ using a probabilistic diffusion model trained on atmospheric 
reanalysis data.
Specifically, the diffusion framework applies a forward process defined by the following stochastic differential equation (SDE):
\begin{equation*}
\label{forward}
    d\mathbf{x}= \mathbf{f}(\mathbf{x},\tau)d\tau + g(\tau)d\mathbf{w}, \quad \tau \in [0,T],
\end{equation*}
where $\mathbf{w}$ is a standard Wiener process,   $\mathbf{f}$ is the drift term, and $g$ is the diffusion term. $\mathbf{f}$ and $g$ are pre-determined so that the forward SDE transforms the target climatological distribution to a standard Gaussian distribution: $p(\mathbf{x}_{\tau=0}) = p_{\theta^*}(\mathbf{X}|\Theta)$, $p(\mathbf{x}_{\tau=T}) = \mathcal{N}(\mathbf{0},\mathrm{I)}$.
The corresponding reverse SDE for generation is given by
\begin{equation*}
\label{reverse}
    d\mathbf{x} = [\mathbf{f}(\mathbf{x},\tau) - g^2(\tau)\nabla_{\mathbf{x}} \log p({\mathbf{x}} )]d\tau + g(\tau)d\bar{\mathbf{w}},
\end{equation*}
where $\bar{\mathbf{w}}$ is a standard Wiener process running backward in time, $\nabla_\mathbf{x}\log p(\mathbf{x})$,  known as the score function, is approximated using a deep neural network $s_\theta$ trained for score matching:

\begin{equation*}
    \nabla_{\mathbf{x}_\tau}\log p(\mathbf{x}_\tau)\approx 
    \argminD_{s_\theta}  \mathbb{E}_{\tau, \mathbf{x}_0}\mathbb{E}_{\mathbf{x}_\tau|\mathbf{x}_0}\|s_\theta(\mathbf{x}_\tau,\tau) - \nabla_{\mathbf{x}_\tau}\log p(\mathbf{x}_\tau|\mathbf{x}_0)\|_2^2
\end{equation*}
We parameterize $s_\theta$ with a U-Net \cite{ronneberger2015u}, and train it to minimize the score matching error via stochastic gradient descent.
We use a second-order probability flow ordinary differential equation solver 
for fast sampling \cite{song2020denoising}. 
Details of data preprocessing, noise scheduling, model architecture, training, and sampling are provided in SI Sec. \textcolor{blue}{B.1}. 

\subsubsection*{$p(\mathbf{X}_{t}|\mathbf{O}_{t})$: enforcing observational constraint}

At the start of assimilation, we combine the learned climatological prior with observational evidence to obtain posterior state estimates, leveraging the pre-trained diffusion model. For notational clarity, we drop the subscript $t_a$ from $\mathbf{X}_{t_a}$ and $\mathbf{O}_{t_a}$ when discussing the diffusion process, and use subscripts to denote diffusion time instead.

To incorporate observational constraints into the generative process, we modify the learned score function to include observational information:


\begin{equation*}
\label{bayes}
    \nabla_{\mathbf{x}_\tau}\log p(\mathbf{x}_\tau|\mathbf{O})
    =\nabla_{\mathbf{x}_\tau}\log p(\mathbf{x}_\tau)+\nabla_{\mathbf{x}_\tau}\log p(\mathbf{O}|\mathbf{x}_\tau)
\end{equation*}
The first term can be quantified by the pre-trained score function $s_\theta$, the second term represents the observational likelihood gradient.
To compute it,
we make two approximation assumptions following the co-paint methodology \cite{zhang2023towards}:
\begin{enumerate}
    \item The observations follow a Gaussian likelihood:
\begin{equation*}
    p(\mathbf{O}|\mathbf{X}):= \mathcal{N}(\mathbf{O}; \Omega(\mathbf{X}), \mathbf{\Sigma}_\mathbf{o})
\end{equation*}
where $\Omega$ represents the observation operator, $\mathbf{\Sigma}_\mathbf{o}$ is the observation error covariance.
\item The final state $\mathbf{X}$ can be approximated from the diffusion state  $\mathbf{x}_\tau$  with uncertainty:
\begin{equation*}
    p(\mathbf{X}|\mathbf{x}_\tau):=\mathcal{N}(\mathbf{X}; F_\tau(\mathbf{x}_\tau), \mathbf{\Sigma}_\mathbf{\tau})
\end{equation*}
where $F_\tau$ is the one-step mapping from $\mathbf{x}_\tau$ to $\mathbf{X}$ for which an analytical form is available \cite{song2020denoising}, and
$\mathbf{\Sigma}_\mathbf{\tau}$ quantifies the uncertainty in this approximation.
\end{enumerate}
Given these assumptions, and noticing that $\mathbf{x}_\tau$ and $\mathbf{O}$ are conditionally independent given $\mathbf{X}$, we can marginalize over $\mathbf{X}$ to obtain the observational likelihood gradient (see SI Sec. \textcolor{blue}{B.3}):

\begin{equation*}
    \nabla_{\mathbf{x}_\tau} \log p(\mathbf{O}|\mathbf{x}_\tau) = (\nabla F_\tau)^\top \cdot (\nabla \Omega)^\top \cdot (\mathbf{\Sigma}_\mathbf{o} + \nabla \Omega \cdot \mathbf{\Sigma}_\tau \cdot \nabla \Omega^\top)^{-1} \cdot (\mathbf{O} - \Omega(F_\tau(\mathbf{x}_\tau)))
\end{equation*}

To practically enforce the observational constraints during sampling, 
we consider direct observations ($\Omega$ as identity mapping) and empirically determine the combinatorial influence of $\mathbf{\Sigma}_\mathbf{o}$ and $\mathbf{\Sigma}_\mathbf{\tau}$,  based on ensemble estimation accuracy for unobserved regions, quantified via Continuous Ranked Probability Score (CRPS, see below).
To further enhance coherence between observed and unobserved regions, we adopt the \textit{time traveling} strategy \cite{lugmayr2022repaint,chao2024learning}: during generation, we periodically return to the previous denoising steps by corrupting the intermediate state estimate.  See SI Sec. \textcolor{blue}{B.3} for details.

\subsubsection*{$p(\mathbf{X}_t|\mathbf{X}_{t-\Delta t},\mathbf{O}_t)$ \& $p(\mathbf{X}_t|\mathbf{X}_{t-\Delta t})$:  enforcing predictive constraints}
To incorporate predictive information into the generative process, we need a forecasting model $\mathbf{M}$ that induces state transition $p(\mathbf{X}_t|\mathbf{X}_{t-\Delta t},\mathbf{O}_t)$  (during assimilation)
and $p(\mathbf{X}_t|\mathbf{X}_{t-\Delta t})$ (during prediction). While various models could serve this purpose, including physics-based approaches, we demonstrate GAP's capabilities using a data-driven forecasting model.


The predictive model $\mathbf{M}$ maps the current atmospheric state to
the next state that is 6 hours ahead. 
This model, which we call ConvCast, is a modified UNet with ConvNeXt  \cite{liu2022convnet} blocks, leveraging depth-wise convolutions with larger kernel sizes for enhanced spatial-temporal pattern recognition.
The model is trained on 39 years of ERA5 reanalysis data (1979-2017) at 1-degree spatial resolution, minimizing a weighted mean squared error loss function aligned with GraphCast's weighting scheme \cite{lam2023learning}. The training process incorporates cosine-annealing learning rate scheduling initialized at 0.001 and utilizes a scheduled drop path with a 0.2 ratio to prevent overfitting. This combination of architectural design and training strategies ensures robust performance in weather prediction tasks while maintaining the computational efficiency needed for operational forecasting applications. SI Sec. \textcolor{blue}{B.2} provides details of this predcitive model.

To enforce predictive constraints, we use Stochastic Differential Equation Editing (SDEdit) \cite{meng2021sdedit}.
Instead of generating state estimates from pure noise, we start the diffusion process from the model prediction $\mathbf{M}(\mathbf{X}_t)$ with carefully calibrated noise. Specifically, we:
\begin{enumerate}
    \item run the forward SDE from $\tau = 0$ to $\tau^*$ with initial condition $\mathbf{x}_{\tau=0}= \mathbf{M}(\mathbf{X}_t)$, where $\tau^*$ tells how much noise is applied to the raw prediction of $\mathbf{M}(\mathbf{X}_t)$.
    \item Run the reverse SDE back from $\tau =\tau^*$ to $0$ to generate the probabilistic estimate for $\mathbf{X}_{t+\Delta t}$, which we denote as $\text{SDEdit}(\mathbf{M}(\mathbf{X}_t), \tau)$. 
\end{enumerate}
The effectiveness of this approach can be understood through error decomposition:
\begin{align*}
\|\mathbf{X}_{t+\Delta t} - \text{SDEdit}(\mathbf{M}(\mathbf{X}_t), \tau)\|_2^2 \leq & \|\mathbf{X}_{t+\Delta t} - \text{SDEdit}(\mathbf{X}_{t+\Delta t}, \tau^*)\|_2^2 + \\&\|\text{SDEdit}(\mathbf{X}_{t+\Delta t}, \tau^*) - \text{SDEdit}(\mathbf{M}(\mathbf{X}_t), \tau^*)\|_2^2
\end{align*}
This inequality decomposes the prediction error into inherent information loss from applying noise to the true state (left), and how differences between the true state and model prediction propagate through the diffusion process (right), with impacts balanced by the critical parameter $\tau^*$:
\begin{enumerate}
    \item For small values of $\tau^*$ and insufficient noise, the distribution remains tightly constrained around $\mathbf{M}(\mathbf{X}_t)$, preserving the systematical biases in $\mathbf{M}$;
    \item For large values of $\tau^*$ and excessive noise, the distribution approaches the unconstrained climatological prior, losing valuable forecast information. 
\end{enumerate}
In practice, we determine the optimal value of $\tau^*$ based on power spectra and probablistic skill. For $p(\mathbf{X}_t|\mathbf{X}_{t-\Delta t},\mathbf{O}_t)$, we apply a similar strategy but include observation inpainting during sampling to ensure consistency with observational constraints. Further details on the SDEdit approach, including theoretical and practical guidance for selecting $\tau^*$, can be found in SI Sec. \textcolor{blue}{B.4} and \textcolor{blue}{B.7}.

\subsubsection*{$p(\mathbf{X}_t|\mathbf{X}_{t-\Delta t},\Theta)$: enforcing external forcing constraints}

To represent how external forcings ($\Theta$)  influence predictions, we capture their impacts through key state variables, primarily focusing on ocean temperatures. Since oceans absorb over 90\% of excess greenhouse gas heat, they contain the most salient climate change signals. Our approach extends the inpainting methodology to ensure atmospheric states remain physically consistent with oceanic conditions, effectively translating ocean climate signals into appropriate weather patterns.

For seasonal prediction, we first project future sea surface temperatures using anomaly persistence, then infer corresponding 2-meter air temperatures, and finally inpaint these constraints throughout simulations (see SI Sec. \textcolor{blue}{B.5} for implementation details). For historical climate simulations, we inpaint sea-air temperatures from ERA5 as boundary forcings. In climate projections, we inpaint sea-air temperatures from CESM2 simulations under different forcing scenarios (SSP1-2.6, SSP3-7.0, SSP5-8.5, see SI Sec. \textcolor{blue}{B.6}).
While this approach intentionally ignores dynamic ocean-atmosphere feedbacks, it enables GAP to bridge weather prediction and climate projection while maintaining physical consistency across timescales without additional training. 
We recognize the limitations of this approach --- particularly the lack of dynamic ocean-atmosphere feedback and reliance on external models to represent climate change signals. Future work will aim to directly represent forcing-response relationships in a coupled modeling framework. 

\subsubsection*{Data}
We train the generative model and the predictive model using hourly atmospheric reanalysis data from ECMWF's ERA5 reanalysis at $1^\circ$ spatial resolution (SI Sec. \textcolor{blue}{A.1}). The training dataset spans from 1979 to 2017 and includes nine key weather variables across 13 standard pressure levels (50-1000 hPa). Initial conditions for model evaluation are derived from ECMWF's High-Resolution Forecast (IFS-HRES analysis, SI Sec. \textcolor{blue}{A.2}), though slight differences exist from the official analysis product due to the absence of additional surface data assimilation steps.
For model evaluation across different timescales, we use several reference datasets:
\begin{itemize}
    \item In data assimilation experiments, we compare performance against ECMWF's High-Resolution Forecast (IFS-HRES analysis) and use observational data from the Integrated Global Radiosonde Archive (IGRA) and Global Hourly Integrated Surface Database (ISD)
    \item For weather forecasting (1-14 days), we benchmark against ECMWF's operational ensemble forecast system (IFS-ENS) as well as recently introduced ML models
    \item For seasonal prediction (1-3 months), we compare with ECMWF's fifth-generation seasonal forecasting system (SEAS5)
    \item For climate simulation and projection experiments, we utilize sea-air temperatures from CESM2 simulations under various shared socioeconomic pathway scenarios (SSP1-2.6, SSP3-7.0, SSP5-8.5).
\end{itemize}


For operational observations, we employ two primary datasets: the Integrated Global Radiosonde Archive (IGRA) and the Global Hourly Integrated Surface Database (ISD). IGRA provides quality-controlled radiosonde observations from over 2,800 globally distributed stations, offering vertical profiles of atmospheric variables since 1905, while ISD integrates hourly surface observations from more than 100 sources, measuring parameters including wind, temperature, pressure, and precipitation (SI Sec. \textcolor{blue}{A.3}). Extensive preprocessing was applied to handle missing values and measurement errors, with specific protocols for near-surface wind components, surface air temperature, pressure, and radiosonde data. All grid-based data underwent first-order conservative regridding following established benchmarking protocols (SI Sec. \textcolor{blue}{A.4}).

\subsection*{Statistical methods}
We implement a comprehensive evaluation framework that spans deterministic, probabilistic, and physical metrics across different spatial and temporal scales.
For deterministic assessment (SI Sec. \textcolor{blue}{C.1}), we use:
\begin{itemize}
    \item The Root Mean Square Error (RMSE) to quantify the magnitude of forecast errors
    \item The Anomaly Correlation Coefficient (ACC) to measure pattern similarity between forecast and observed anomalies
\end{itemize}
For probabilistic evaluation of ensemble forecasts (SI Sec. \textcolor{blue}{C.2}), we employ:
\begin{itemize}
    \item The Continuous Ranked Probability Score (CRPS) to assess the overall quality of probabilistic forecasts
    \item The Spread-Skill Ratio (SSR) to evaluate ensemble reliability by comparing ensemble spread with forecast error
\end{itemize}
To assess physical consistency and scale-dependent behavior, we analyze
\begin{itemize}
    \item spectral characteristics using zonal Fourier analysis and global spherical harmonic decomposition (SI Sec. \textcolor{blue}{C.3})
    \item statistical distribution matching using the Kolmogorov-Smirnov test at individual grid points (SI Sec. \textcolor{blue}{C.4})
    \item Multi-variable consistency through derived fields including wind speed, mass streamfunction, geostrophic/ageostrophic wind separation, humidity parameters, and water content (SI Sec. \textcolor{blue}{C.5})
\end{itemize}

For high-impact weather and climate phenomena, we implement specialized detection algorithms for tropical cyclones (based on pressure minima and warm-core structure, SI Sec. \textcolor{blue}{C.6.1}), atmospheric blocking (using geopotential height anomaly persistence, SI Sec. \textcolor{blue}{C.6.2}) and major oscillation indices (Arctic and Southern Oscillation, SI Sec. \textcolor{blue}{C.7}).

\section*{Data availability}

All datasets used in this study are publicly available through their respective repositories. The ERA5 reanalysis data were obtained from the Copernicus Climate Change Service's Climate Data Store (CDS) (\href{https://cds.climate.copernicus.eu}{https://cds.climate.copernicus.eu}). The ECMWF ensemble (ENS) and high-resolution (HRES) forecast datasets were accessed from (\href{https://apps.ecmwf.int/datasets/data/tigge/}{https://apps.ecmwf.int/datasets/data/tigge/}). For the quickest access to these datasets, the WeatherBench2 data archive provides an efficient alternative (\href{https://console.cloud.google.com/storage/browser/weatherbench2}{https://console.cloud.google.com/storage/browser/weatherbench2}). The Integrated Surface Database (ISD) was obtained from the National Centre for Environment Information (NCEI) (\href{https://www.ncei.noaa.gov/products/land-based-station/integrated-surface-database}{https://www.ncei.noaa.gov/products/land-based-station/integrated-surface-database}), while the Integrated Global Radiosonde Archive (IGRA) sounding data were accessed through the NCEI Weather Balloon Database (\href{https://www.ncei.noaa.gov/products/weather-balloon/integrated-global-radiosonde-archive}{https://www.ncei.noaa.gov/products/weather-balloon/integrated-global-radiosonde-archive}). Additional model datasets include the North American Multi-Model Ensemble (NMME) (\href{http://iridl.ldeo.columbia.edu/SOURCES/.Models/.NMME}{http://iridl.ldeo.columbia.edu/SOURCES/.Models/.NMME}), SEAS5 from the Copernicus CDS (\href{https://cds.climate.copernicus.eu/datasets/seasonal-monthly-pressure-levels}{https://cds.climate.copernicus.eu/datasets/seasonal-monthly-pressure-levels}), and CESM2 from the Copernicus CDS CMIP6 Archive (\href{https://cds.climate.copernicus.eu/datasets/projections-cmip6}{https://cds.climate.copernicus.eu/datasets/projections-cmip6}). The Supplementary Information materials are available from the corresponding author upon request.

\section*{Code availability}

The code implementing GAP will be available on the GitHub repository. Model configurations, analysis scripts, and data files used in this study will be made publicly available upon acceptance of this work.

\section*{Acknowledgements}
This research is supported by the National Key Research and Development Program of China (Grant Numbers 2023YFC3007700 and 2023YFC3007705).
N.B. acknowledges funding from the Volkswagen Foundation and the European Union's Horizon Europe research and innovation program under Grant Agreement No. 101137601 (ClimTip).
We thank:
Wenyu Zhou, Peishi Jiang, Jian Lu from Pacific Northwest National Laboratory,
Gang Chen from University of California, Los Angeles,
Yi Deng from Georgia Institute of Technology,
Ming Pan from University of California, San Diego,
Y. Q. Sun from University of Chicago,
Peili Wu from U.K. Meteorological Office,
Yu Huang, Philipp Hess, and Michael Aich from Technical University of Munich,
Xin Li from Chinese Academy of Sciences,  Institute of Tibetan Plateau Research,
Juanjuan Liu, Bin Wang, Li Dong, Guiwan Chen, and Will Yao from Chinese Academy of Sciences, Institute of Atmospheric Physics,
Yuling Jiao, Cheng Yuan from Wuhan University,
Qingyun Duan, Wentao Li from Hohai University,
Wei Gong from Beijing Normal University,
Wenbin Chen, Yutao Fan, Yang Liu from Nanjing University,
Sencan Sun from Tsinghua University, 
for their insightful discussions and valuable feedback on our work. We also thank the European Centre for Medium-Range Weather Forecasts (ECMWF) for sharing the reanalysis dataset and Weatherbench2 for giving access to reanalysis and model data.


\bibliography{sn-bibliography}

\end{document}